\documentclass[sigconf]{acmart}
\usepackage[a-1b]{pdfx}

\copyrightyear{2023}
\acmYear{2023}
\setcopyright{rightsretained}

\acmConference[SIGIR '23]{Proceedings of the 46th International ACM SIGIR Conference on Research and Development in Information Retrieval}{July 23--27, 2023}{Taipei, Taiwan.}
\acmBooktitle{Proceedings of the 46th Int'l ACM SIGIR Conference on Research and Development in Information Retrieval (SIGIR '23), July 23--27, 2023, Taipei, Taiwan}

\makeatother

\usepackage{enumitem}
\newlist{inlinelist}{enumerate*}{1}
\setlist*[inlinelist,1]{%
  label=(\roman*),
}
\usepackage{subfigure}
\usepackage{xcolor}
\usepackage{multirow}
\usepackage{xspace}
\usepackage{todonotes}
\usepackage{array}

\newcommand{\drframework}{DEDR\xspace}
\newcommand{\mmfid}{MM-FiD\xspace}
\usepackage{tablefootnote}

\title{A Symmetric Dual Encoding Dense Retrieval Framework for Knowledge-Intensive Visual Question Answering}

\author{Alireza Salemi}
\affiliation{\institution{University of Massachusetts Amherst}
\country{United States}}
\email{asalemi@cs.umass.edu}

\author{Juan Altmayer Pizzorno}
\affiliation{\institution{University of Massachusetts Amherst}
\country{United States}}
\email{jpizzorno@cs.umass.edu}

\author{Hamed Zamani}
\affiliation{\institution{University of Massachusetts Amherst}
\country{United States}}
\email{zamani@cs.umass.edu}

\copyrightyear{2023}
\acmYear{2023}
\setcopyright{acmlicensed}
\acmConference[SIGIR '23] {Proceedings of the 46th International ACM SIGIR Conference on Research and Development in Information Retrieval}{July 23--27, 2023}{Taipei, Taiwan.}
\acmBooktitle{Proceedings of the 46th International ACM SIGIR Conference on Research and Development in Information Retrieval (SIGIR '23), July 23--27, 2023, Taipei, Taiwan}
\acmPrice{15.00}
\acmISBN{978-1-4503-9408-6/23/07}
\acmDOI{10.1145/3539618.3591629}

\begin{document}


\begin{abstract}
Knowledge-Intensive Visual Question Answering (KI-VQA) refers to answering a question about an image whose answer does not lie in the image. This paper presents a new pipeline for KI-VQA tasks, consisting of a retriever and a reader. First, we introduce \drframework, a symmetric dual encoding dense retrieval framework in which documents and queries are encoded into a shared embedding space using uni-modal (textual) and multi-modal encoders. We introduce an iterative knowledge distillation approach that bridges the gap between the representation spaces in these two encoders. Extensive evaluation on two well-established KI-VQA datasets, i.e., OK-VQA and FVQA, suggests that \drframework outperforms state-of-the-art baselines by 11.6\% and 30.9\% on OK-VQA and FVQA, respectively.

Utilizing the passages retrieved by \drframework, we further introduce \mmfid, an encoder-decoder multi-modal fusion-in-decoder model, for generating a textual answer for KI-VQA tasks. \mmfid encodes the question, the image, and each retrieved passage separately and uses all passages jointly in its decoder. Compared to competitive baselines in the literature, this approach leads to 5.5\% and 8.5\% improvements in terms of question answering accuracy on OK-VQA and FVQA, respectively.
\end{abstract}



\maketitle

\section{Introduction}
\begin{figure}[t]
    \centering
    \includegraphics[width=.8\linewidth]{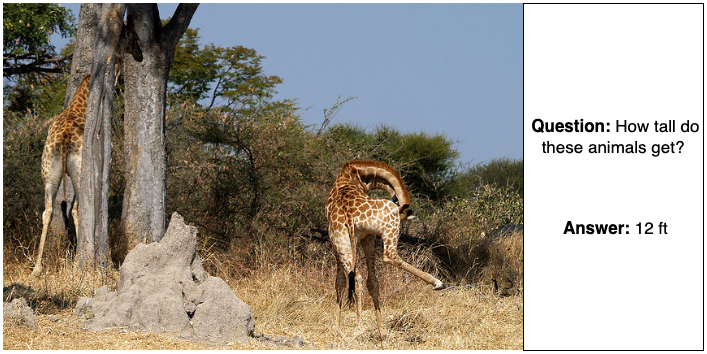}
    \vspace{-0.3cm}
    \caption{An example KI-VQA question. Answering these question requires external knowledge. \\     {\normalfont\footnotesize Image \copyright~ zrim,
        \url{https://www.flickr.com/photos/zrimshots/2788695458}}} 
    \label{fig:example}
    \vspace{-0.3cm}
\end{figure}

Knowledge-intensive visual question answering\footnote{This task is also referred to as outside-knowledge visual question answering (OK-VQA) in the literature \cite{okvqa}. OK-VQA is also the name of a dataset used in this paper. To avoid confusion, we use ``KI-VQA'' to refer to the task.} (KI-VQA) is a variant of visual question answering tasks whose questions cannot be answered using the image alone. Therefore, accessing external knowledge sources is necessary to answer these questions. KI-VQA has a large number of real-world applications. Imagine customers of e-commerce websites taking a photo of a product or a part of a product and asking a question about it. In the context of education, students can ask a question about an image in their textbook. Users can take a photo of a visual sign or a piece of art and ask questions about its meaning or history. These are just a few examples of KI-VQA applications. Figure~\ref{fig:example} shows an example of KI-VQA tasks: the image is sufficient to identify the ``animals'' as giraffes, and likely also the the subspecies, but not to answer how tall they get.

The majority of prior work on KI-VQA, such as \cite{conceptbert, unifer, rvl, krisp,mavex}, assumes that the external knowledge can be obtained from a structured knowledge base. However, a high-quality and complete knowledge base may not be available for some domains \cite{ABUSALIH2021103076}. Besides, maintaining knowledge bases with up-to-date information is challenging \cite{tang-etal-2019-learning}. To prevent these issues, following \citet{okvqa-passage-retrieval}, we take an alternative approach to KI-VQA: using a large text corpus as the external knowledge source. In this setting, a two-stage pipeline for KI-VQA systems is to first retrieve a list of passages for a given question-image pair and then process the retrieved passages to generate an answer.\footnote{This is similar to the retriever and reader stages in open-domain question answering tasks \cite{reading-wikipedia}.} This pipeline is depicted in Figure~\ref{fig:okvqa-pipeline}.

The effective performance of dense retrieval models in various information retrieval tasks \cite{dpr,ADORE,cl-drd} and their extension flexibility to multi-modal\footnote{In this paper, multi-modality refers to the combination of text and image.} input have motivated us to focus on dense retrieval for implementing the first stage of the KI-VQA pipeline (i.e., passage retrieval). A property of this retrieval task is that it deals with asymmetric input modalities: the user information need is multi-modal (question-image pair) while the information items (passages) are uni-modal. As a result of this property, \citet{okvqa-passage-retrieval} recently showed that a KI-VQA dense retrieval model that uses a multi-modal encoder for representing the question-image pair and a text encoder for representing the passages in the collection leads to state-of-the-art passage retrieval performance. We argue that using such an asymmetric bi-encoder architecture is sub-optimal, since the encoders produce outputs in different semantic spaces and fine-tuning the encoders cannot always close this gap. We first study two alternatives for developing symmetric dense retrieval models:\footnote{Symmetric dense retrieval refers to a bi-encoder architecture with shared parameters.} (1) producing a textual representation of the image and using a symmetric uni-modal bi-encoder architecture for dense retrieval, and (2) converting passages to a multi-modal input format and using a symmetric multi-modal bi-encoder architecture. We observe that both alternatives suffer from information loss, but also that they produce complementary representations. 
This observation motivates us to not only combine these two encodings, but also transfer knowledge between them. In more detail, we propose an iterative knowledge distillation approach to transfer knowledge between these two alternative symmetric dense retrieval models. The proposed symmetric dual encoding approach leads to 11.6\% and 30.9\% MRR improvements compared to the state-of-the-art baseline on OK-VQA \cite{okvqa} and FVQA \cite{fvqa} test sets, respectively.

For the second stage of the pipeline, unlike much prior work on answer span detection for KI-VQA \cite{conceptbert, unifer, mavex, rvl, krisp} (i.e., answer extraction from the retrieved passages), we focus on retrieval-augmented autoregressive answer generation. We propose MM-FiD, a simple yet effective extension of the Fusion-in-Decoder (FiD) ~\cite{fid} architecture to multi-modal input. FiD is a retrieval-augmented text generation model that has recently shown effective performance in question answering tasks \cite{fid}. MM-FiD uses a multi-modal encoder to represent the question, the image, and the retrieved passages and uses a uni-modal decoder that generates an answer and is trained using the maximum likelihood objective. Extensive experiments on both OK-VQA and FVQA datasets demonstrate that MM-FiD significantly outperforms alternative approaches for answer generation. It also performs better than answer span detection baselines. In more detail, our end-to-end pipeline achieves 5.5\% and 8.5\% improvement compared to the baselines on OK-VQA and FVQA question answering tasks, respectively. We open-source our code and release our learned model parameters for research purposes\footnote{\url{https://github.com/alirezasalemi7/DEDR-MM-FiD}}.

\section{Related Work}

\subsubsection*{\textbf{(Multi-Modal) Dense Retrieval}}

Using dense vectors for retrieving textual documents related to a textual query has been studied since the emergence of Latent Semantic Analysis \cite{lsi}. However, dense retrievers' performance remained inferior to that of sparse retrievers like BM25 until \citet{dpr}'s Dense Passage Retriever (DPR), which uses the \texttt{[CLS]} token output by BERT~\cite{bert}, a pre-trained language model. While many dense retrievers only use a single vector to represent the query and the document \cite{xiong2021approximate, dpr, qu-etal-2021-rocketqa}, using multiple vectors per document and query has been also studied \cite{colbertv1, colbertv2, Humeau2020Poly-encoders:, 10.1145/3397271.3401093, gao-etal-2020-modularized}. 


Multi-modal dense retrieval has recently been investigated in different forms: (1) uni-modal query and multi-modal documents \cite{manymodal,talmor2021multimodalqa, https://doi.org/10.48550/arxiv.2209.00179}, (2) multi-modal query and uni-modal documents \cite{okvqa-passage-retrieval}, (3) multi-modal query and multi-modal documents \cite{singh-etal-2021-mimoqa}, and (4) uni-modal query and uni-modal documents with queries and documents from different modalities, i.e., cross-modal retrieval \cite{align, clip}.

In this work, we focus on the second case, where the query is multi-modal while the documents only contain text.
\citet{okvqa-passage-retrieval} utilized an asymmetric bi-encoder with LXMERT~\cite{lxmert}, a pre-trained vision-language model based on BERT~\cite{bert} for encoding queries, and BERT itself for encoding documents. As we show, such an asymmetric architecture is sub-optimal; utilizing different encoders creates a semantic ``gap'' in the embedding space and fine-tuning cannot easily overcome the issue. We instead propose a new symmetric dual encoding framework that addresses this issue.

\subsubsection*{\textbf{Knowledge Distillation for Dense Passage Retrieval}}

Due to the vast number of learnable parameters in dense passage retrievers, sometimes available datasets are insufficient to train them \cite{cl-drd}. Consequently, knowledge distillation, in which a teacher model provides labels for a student model, has become a standard approach for training dense retrieval models and has shown compelling outcomes \cite{10.1145/3404835.3462891, lin-etal-2021-batch}. Existing work in this area often uses cross-encoder rerankers that input both query and document as teacher models for dense retrieval models with a bi-encoder architecture \cite{ren-etal-2021-rocketqav2}. Another approach is to distill knowledge from multi-vector dense retrieval models, such as ColBERT \cite{colbertv1} to single-vector dense retrievers \cite{ADORE}.

In our experiments, we did not find knowledge distillation from cross-encoder rerankers helpful for KI-VQA datasets. Thus, we introduce a novel approach: iterative knowledge distillation for dual encoding, in which knowledge distillation happens iteratively between two bi-encoders that use different modalities in their inputs. Accordingly, each model learns the perspective of the other, adjusting their representation spaces for more effective dense retrieval. 

\subsubsection*{\textbf{Knowledge-Intensive Visual Question Answering}}

\emph{Knowledge-intensive} refers to a category of retrieval-enhanced machine learning problems \cite{reml} whose inputs are not sufficient to produce the output and external information should be provided. 
KILT \cite{kilt} is a benchmark for natural language knowledge-intensive tasks such as open-domain question answering, fact-checking, entity-linking, slot-filling, and knowledge-based dialogue. All the mentioned tasks are text-only knowledge-intensive tasks.
To the best of our knowledge, there is no unified benchmark on multi-modal knowledge-intensive tasks, which is relatively less explored. Therefore, this paper focuses on knowledge-intensive visual question answering. 

Visual question answering (VQA) is a multi-modal question answering task whose goal is to answer a natural language question about an image \cite{VQA}. VQA is primarily designed to measure the ability of models in representing and comprehending multi-modal data. Therefore, questions in VQA are often related to visual features (e.g., color or shape of the objects) and sometimes require commonsense knowledge. In other words, a human can answer VQA questions by just looking at the image, without accessing external information. Given this formulation, VQA has limited real-world use cases. In contrast to VQA,
knowledge-intensive visual question answering is the task of answering a question about an image that needs an external piece of information not available in the image to answer the questions. 
Fact-based Visual Question Answering (FVQA) \cite{fvqa} is a visual question-answering dataset in which answering the questions about an image needs the model to consider a relevant fact. Alternatively, outside-knowledge visual question answering (OK-VQA) \cite{okvqa} is a dataset similar to  FVQA, but the required knowledge is not limited by facts. 


Current approaches for OK-VQA utilize different strategies to solve the problem: some rely on implicit knowledge stored in language or vision-language models to answer the questions without using any external source of knowledge \cite{cbm, pica}, while others use external knowledge sources for this purpose in addition to implicit knowledge \cite{kat, krisp, conceptbert,mavex,unifer}. Using OCR and dense object labels can also be effective for this task \cite{https://doi.org/10.48550/arxiv.2210.03809, 9879942}, but is beyond the scope of this paper.
In order to use an explicit knowledge source, it is necessary to design a \emph{retriever} that retrieves a small set of relevant passages to the image and question \cite{okvqa-passage-retrieval} and a \emph{reader} that selects or generates the response from the retrieved passages \cite{kat, unifer, vrr}. This paper proposes effective solutions for both of these steps.


\begin{figure}
    \centering
    \includegraphics[width=\linewidth]{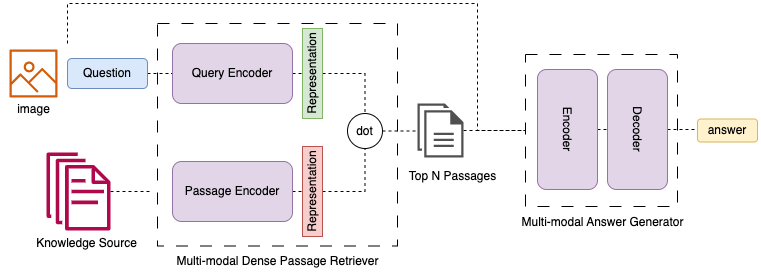}
    \vspace{-0.6cm}
    \caption{A pipeline for KI-VQA tasks that retrieves unstructured text in response to the input question-image pair and uses the retrieved passages as supporting documents to generate textual answer.} 
    \label{fig:okvqa-pipeline}
    \vspace{-0.5cm}
\end{figure}



\section{Problem Statement}
In knowledge-intensive visual question answering, a user asks a natural language question about an image, which requires access to external information. In other words, the answer to the question does not exist in the image, which necessitates the utilization of external resources. See Figure~\ref{fig:example} for an example of KI-VQA tasks. These resources can be in many different forms, from structured and semi-structured knowledge bases to unstructured text retrieved from the web. In this paper, we consider a scenario where the answer should be retrieved and extracted from a collection of unstructured natural language passages. In the following, we specify more formally the KI-VQA task studied in this paper.

Let $T = \{(Q_1, I_1, A_1, R_1), (Q_2, I_2, A_2, R_2), \cdots, (Q_n, I_n, A_n, R_n)\}$ denote the training set for a KI-VQA task. Each training instance consists of a natural language question $Q_i$, an image $I_i$, a set of short textual answers $A_i$, and a set of relevant passages $R_i$. That means each question \emph{may} have multiple answers in the training set (i.e., $|A_i| \geq 1$), which are often semantically the same but syntactically different. Similarly, there may exist multiple passages that are relevant to the question (i.e., $|R_i| \geq 1$). All relevant passages are selected and annotated from a large-scale collection $\mathcal{C}$. Therefore, $R_i \subseteq \mathcal{C}: \forall 1 \leq i \leq n$. We study the following two related tasks:

\medskip

\noindent \textbf{Passage Retrieval for KI-VQA}: the retrieval task is to use the training set $T$ to train a retriever that retrieves relevant passages from the collection $\mathcal{C}$ for a given question-image pair $(Q, I)$. 

\medskip

\noindent \textbf{Retrieval-Augmented Answer Generation for KI-VQA}: the retrieval-augmented answer generation task is to generate a short textual answer for any unseen question-image pair $(Q, I)$ by having access to the collection $\mathcal{C}$. Therefore, models in this task naturally retrieve passages from $\mathcal{C}$ and utilize them for generating an answer.

\medskip

We first propose a symmetric dual encoding architecture for dense retrieval in KI-VQA and then introduce a multi-modal fusion-in-decoder model as a retrieval-enhanced answer generation approach.

\section{\drframework: Dual Encoding Dense Retriever Framework}
Figure~\ref{fig:okvqa-pipeline} depicts a pipeline for knowledge-intensive visual question answering tasks. As shown in the pipeline, the input to the dense retrieval model is asymmetric -- query encoder takes multi-modal input (i.e., a question and an image), while the passage encoder takes a uni-modal text input (i.e., a passage from $\mathcal{C}$). This asymmetric property in the input modalities makes it challenging to design an effective symmetric dense retrieval model. This is why the current state-of-the-art dense retrieval model proposed by \citet{okvqa-passage-retrieval} uses an asymmetric architecture, where a pre-trained multi-modal language model (i.e., LXMERT \cite{lxmert}) is used for query encoding and a pre-trained uni-modal language model (i.e., BERT \cite{bert}) is used for document encoding.
Since such asymmetric architectures start from fundamentally different embedding spaces, they suffer from slow convergence speed and sub-optimal dense retrieval performance. Conversely, extensive research on dense retrieval for uni-modal data (textual queries and documents) suggests that symmetric architectures lead to significantly better performance. State-of-the-art dense passage retrieval models, such as TAS-B \cite{10.1145/3404835.3462891}, ColBERT \cite{colbertv1, colbertv2}, RocketQA \cite{ren-etal-2021-rocketqav2,qu-etal-2021-rocketqa}, and CLDRD \cite{cl-drd}, use symmetric architectures. Motivated by this observation, our goal is to learn a \emph{symmetric dense retrieval model} for KI-VQA tasks. 

To this aim, we study two alternative solutions. First, we convert all model inputs to a uni-modal textual form and then use uni-modal language models for both query and document encoding (Section \ref{sec:unified-text}). Second, we convert all inputs to the same multi-modal (text and image) form and then use multi-modal language models for both encoders (Section \ref{sec:unified-multimodal}). We hypothesize that these two models learn complementary representations for the following reasons: (1) they take different input formats, and (2) the pre-training process and data in uni-modal and multi-modal language models are different. Our experimental results also validate this hypothesis (see Section \ref{sec:passage-ret-results}). Following this observation, we propose an iterative knowledge distillation approach that alternates between these two encoding approaches as teacher and student models. Finally, by combining these two encoding approaches \drframework learns a symmetric dual uni-modal/multi-modal encoder for both queries and documents.

\begin{figure*}
    \centering
    \includegraphics[trim="0cm 1cm 2cm 0cm",clip,width=\textwidth]{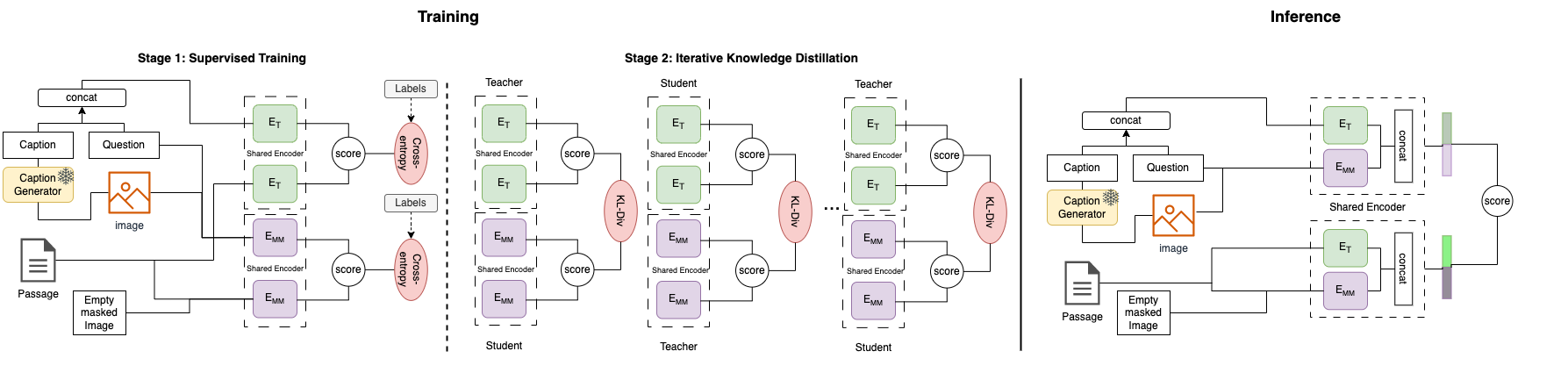}
    \vspace{-0.5cm}
    \caption{The training and inference procedure in the \drframework framework. \drframework first trains uni-modal and multi-modal encoders in isolation (left), then uses iterative knowledge distillation to adjust both representation spaces (middle). At inference, \drframework uses the aggregation of both encodings to construct a symmetric dual encoding dense retriever (right).}
    \label{fig:dedr}
    \vspace{-0.4cm}
\end{figure*}

\subsection{Unified Uni-Modal Encoding}
\label{sec:unified-text}
In order to use a shared uni-modal (textual) encoder for representing both queries and passages, we need to convert the image in the query to text. This model can be formulated as follows:
\begin{equation}
    S_T((Q, I), P) = E_T (\text{concat}(Q, \phi_{I \rightarrow T}(I))) \cdot E_T(P)
    \label{eq:text_encoder}
\end{equation}
where $\cdot$ denotes dot product between two vectors, $E_T$ is a uni-modal text encoder, and $\phi_{I \rightarrow T}$ is a modality converting module that takes an image and produces a textual description for it. 

There are several approaches to implement the modality converter $\phi_{I \rightarrow T}$. One approach is to generate the name of objects that are in the image using object detection approaches. We take an alternative approach by using image captioning objectives to train the modality converter model $\phi_{I \rightarrow T}$. The reason is that image captioning approaches produce open-ended descriptions of images, as opposed to predefined categories in object detection models. In addition, collecting training data for image captioning models is cheap, given the availability of large-scale images with captions on the web. In more detail, we use the ExpansionNet v2 \cite{expansionv2} architecture to implement $\phi_{I \rightarrow T}$. Expansion V2 is an encoder-decoder architecture designed on top of the Swin-Transformer \cite{swin-transformer} extended by Block Dynamic and Static Expansion \cite{expansionv2} and multi-head attention \cite{transformer}. The model is first pre-trained using images and captions from the Microsoft COCO dataset \cite{mscoco}. Then, the self-critical optimization \cite{Rennie2016SelfCriticalST} is performed to complete the model's training. Once the model is trained for producing textual descriptions of images, we freeze the ExpansionNet~v2's parameters and only optimize the text encoder $E_T$. This substantially reduces the number of parameters that need to be learned using the KI-VQA training set.

For implementing the text encoder $E_T$, there are numerous language models available. In our experiments, we use BERT-base \cite{bert} and the representation associated with the \texttt{[CLS]} token is considered as the output of the encoder $E_T$. Note that in Equation~\eqref{eq:text_encoder}, query and passage encoders are the same and they use shared parameters, guaranteeing a symmetric architecture for dense retrieval.


\subsection{Unified Multi-Modal Encoding}
\label{sec:unified-multimodal}
Even though using a multi-modal encoder for both query and passage encoding seems straightforward, most multi-modal language models do not accept text-only inputs. Therefore, we need to develop a technique to fill this modality gap. Our unified multi-modal encoding approach can be formulated as follows:
\begin{equation}
    S_{MM}((Q, I), P) = E_{MM}(Q, I) \cdot E_{MM}(P, I_\texttt{[MASKED]})
    \label{eq:multimodal_encoder}
\end{equation}
where $E_{MM}$ is a pre-trained multi-modal encoder that represents a pair of text and image as input. To address the modality gap between the query and passage sides, we use a multi-modal language model that has used the masking technique in the visual side during pre-training. In more detail, we use LXMERT \cite{lxmert} that uses Faster R-CNN \cite{faster-r-cnn} to recognize 36 objects in the given image and generate their representations and bounding boxes. Then, this information about objects in addition to the text tokens are fed to a dual-encoder transformer network with a cross-modality encoder on top. Finally, the \texttt{[CLS]} token is used to represent the whole input. Since LXMERT has been pre-trained with different pre-training objectives, including Masked Object Prediction, it is a perfect fit for our unified multi-modal representation learning. 

In order to overcome the aforementioned problem with encoding textual only data for passage representation, we propose a simple yet effective technique that we call Passage Expansion using Masked Image Representation (PEMIR). In this technique, we feed a passage to LXMERT as the textual input and zero (masked) as the visual input with bounding boxes of $[0.0, 0.0, 1.0, 1.0]$. This visual input means 36 masked objects with bounding boxes of the whole image. Intuitively, we ask the model to generate a representation for the input passage while trying to generate the best visual object representations based on the textual input data. Thus, this is roughly equivalent to expanding the passage with image representation based on the passage content. 

Using this approach, we can generate the representation for queries and documents using the same multi-modal encoder with shared parameters. This is beneficial because it helps the dual encoder architecture start from the same shared embedding space. Additionally, we can use only a single encoder as both query and document encoder, which results in decreasing the number of parameters of the model. 

\subsection{Dual Encoding Optimization via Iterative Knowledge Distillation}

\begin{figure}
    \centering
    \includegraphics[width=\linewidth]{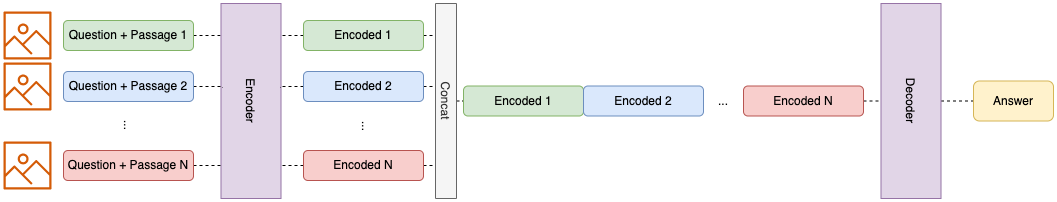}
    \vspace{-0.5cm}
    \caption{The architecture of \mmfid. It uses multi-modal encoder to encode each question-image-passage triplet separately and then concatenates their encodings as input to the decoder for knowledge aggregation and answer generation.}
    \label{fig:multi-fid}
    \vspace{-0.5cm}
\end{figure}

$E_T$ and $E_{MM}$ represent the KI-VQA inputs from different perspectives; $E_T$ only considers the textual representation of the inputs, while $E_{MM}$ considers their multi-modal representations. These models also use unique pre-training data and objectives. Hence, we hypothesize that these two representation learning models provide complementary information. Our empirical analysis validates this hypothesis, see Figure~\ref{fig:mrr-diff-lxmert-captbert}. Given this observation, we propose to use a knowledge distillation approach to improve both of these models. 

\subsubsection*{\textbf{Isolated Training of Encoders.}} We first optimize the parameters of these two models separately using the retrieval training set for each KI-VQA task. Following DPR \cite{dpr}, we use a contrastive loss function based on cross-entropy to train the models:
\begin{equation}
    \label{eq:contrastive-loss}
    L_{\text{isolated}} = - \log{\frac{e^{S_{X}((Q,I),P_{pos})}}{e^{S_{X}((Q,I),P_{pos})} + \sum_{P' \in \mathbf{P_{neg}}} e^{S_{X}((Q,I),P')}}}
\end{equation}
where $X \in \{T, MM\}$ is the encoding approach used to generate scores, and $P_{pos}$ is a positive (relevant) passage and $\mathbf{P_{neg}}$ is a set of negative passages for the question-image pair $(Q,I)$. To construct $\mathbf{P_{neg}}$, we use a hard negative sampling approach used in \cite{okvqa-passage-retrieval} in addition to in-batch negatives -- in which all the positive and negative documents of other queries in the training batch are considered as negative documents to the query. 

\subsubsection*{\textbf{Iterative Knowledge Distillation among Encoders.}}
In order to adjust the representations space in $E_T$ and $E_{MM}$ and improve their generalization, we design an iterative knowledge distillation approach. In this method, we first use the more effective encoder, based on the performance of the validation set,  as the teacher and the other encoder as the student. Then, we train the student using the scores provided by the teacher. We use the following listwise KL-divergence loss function:
\begin{equation}
    \label{eq:distilation}
    L_{\text{IKD}} = - \sum_{P' \in \mathbf{P_{neg}} \cup \{P_{pos}\}} S'_{Y}((Q,I),P')\log{\frac{S'_{X}((Q,I),P')}{S'_{Y}((Q,I),P')}}
\end{equation}
%
%
where $X$ and $Y$ respectively denote the student and teacher model. $S'((Q,I),P')$ is the normalized score of $S((Q,I),P')$ for $P' \in \mathbf{P_{neg}} \cup \{P_{pos}\}$  generated by the student or teacher model using the softmax function. Similar to Equation \eqref{eq:contrastive-loss}, $P_{pos}$ is a positive passage, and the same method is used for negative sampling. We continue the training of the student using the teacher's scores until a stopping criterion is met: either the computing budget finishes (i.e., it reaches the maximum number of epochs set in the experiment) or early stopping based on validation performance.

In the next round of distillation, we swap the teacher and the student. In other words, the student of the previous round acts as the teacher in this round to provide scores for the previous teacher's training in this round. This iterative approach to knowledge distillation is helpful because, in each round, the student model learns the perspective of the teacher model in scoring documents, especially when these two models rely on two different embedding spaces to generate scores. We continue this iterative distillation process and use early stopping based on validation performance to terminate. 

\subsection{Retrieval Inference using Dual Encoding}

As we mentioned earlier, $E_T$ and $E_{MM}$, introduced in previous sections, retrieve passages in response to a multi-modal query using separate embedding spaces. The former focuses on textual embedding space, while the latter encodes queries and passages into a multi-modal embedding space. An idea to combine these two models in a single embedding space is to concatenate each model's representation for its inputs together. If each of these encoders produce a $d$-dimensional encoding for each input, the final embedding space consists of $2d$ dimensions.  

There are two methods to use this combined embedding space: (1) we can use the concatenated representation of the models to train a new ranker from scratch using the loss function in Equation \ref{eq:contrastive-loss}, and (2) we can use the best rankers of each type after knowledge distillation and combine their representations without further training to generate representations for queries and passages. The latter does not need training because each model has been trained previously. We just combine their representations to index the embeddings using Faiss \cite{faiss} and search the index to retrieve passages. 

\section{Multi-Modal Fusion-in-Decoder}

Fusion-in-Decoder (FiD) \cite{fid} is a state-of-the-art generative encoder-decoder model based on T5 \cite{t5}. It is intended to aggregate knowledge across multiple passages and generate a single response based on them. It has produced strong performance on a wide range of knowledge-intensive language tasks (KILTs). The current state-of-the-art model on six benchmarks from the KILT leaderboard\footnote{\url{https://eval.ai/web/challenges/challenge-page/689/leaderboard/}} is based on a light variant of FiD \cite{fidlight}. We extend FiD architecture to multi-modal data and propose \mmfid.

To formalize the task, for a query consisting of an image $I$ and a question $Q$, we assume a set $P=\{p_1, ..., p_n\}$ including $n$ supporting passages is provided (e.g., by a retrieval model). The goal of the multi-modal fusion-in-decoder (\mmfid) is to generate a textual answer to $Q$ about the $I$ by considering all passages in the $P$. 

We use VL-T5 \cite{vlt5} architecture, a multi-modal text generative visual-language model pre-trained with different text generation tasks, as a start point to design multi-modal fusion-in-decoder architecture. VL-T5 takes a piece of text and image objects' features detected by Faster R-CNN \cite{faster-r-cnn} as input and generates a piece of text as output. The simplest solution to the mentioned problem formulation in this section is to feed VL-T5 with the concatenation of the question, image, and each passage to generate an answer based on each supporting document; however, the aggregation and reduction of the generated answers for each document are challenging because the model might generate a different answer for each document. Another approach to solving the problem is to concatenate the question, image, and all documents and feed it to VL-T5 to generate a single answer based on them; this approach suffers from two shortcomings: (1) concatenating all passages results in a long sequence, which decreases the speed of the model and increases the memory consumption and may also reach the maximum token limit, and (2) concatenation of different passages together makes it hard for the model to consider the context of each passage because the passage set $P$ can be about different unrelated subjects.

\mmfid uses an alternative approach, shown in Figure \ref{fig:multi-fid}. In this architecture, the question and image are concatenated with each document in the supporting set independently and fed to the encoder of VL-T5 to be encoded. Then, the encoded representation of each question, image, and each passage is concatenated together to be fed to the decoder of VL-T5. In this approach, the image, question, and each document are encoded separately, which helps the model decrease memory use and consider the context better than the two other alternatives mentioned above. Additionally, concatenating the encoded representations of all passages at the decoder helps the model consider the information in all documents for generating a single answer to the question about the image.

In order to train the \mmfid for text generation, we use the cross-entropy loss function using the following formulation:
\begin{equation*}
    L_{\text{mm-fid}} = -\sum_k \log P(y_k| y_{i < k}, \{I,Q,p_1\},\{I,Q,p_2\},...,\{I,Q,p_n\})
\end{equation*}
where $y_k$ is $k$\textsuperscript{th} output token, $I$ is the image, $Q$ is the question, and $p_i$ is the $i$\textsuperscript{th} supporting (e.g., retrieved) passage.

\section{Experiments}


\subsection{\textbf{Datasets}}



\noindent \textbf{Outside-Knowledge Visual Question Answering (OK-VQA)} \cite{okvqa}: This dataset consists of triplets, including an image, a question about the image, and an answer to the mentioned question. Answering most of the questions in this dataset needs a piece of information that is not provided in the image. Therefore, accessing an external source of information is required for this task. A retrieval dataset based on a Wikipedia dump\footnote{This Wikipedia collection is available at \url{https://ciir.cs.umass.edu/downloads/ORConvQA/all_blocks.txt.gz}.} with 11 million passages was later constructed by \citet{okvqa-passage-retrieval}, which we use to train and evaluate our retrievers. This dataset contains 9009 questions for training, 2523 questions for validation, and 2523 for testing \cite{qu-etal-2021-rocketqa}. We also use original OK-VQA dataset to evaluate the performance of our end-to-end retrieval and answer generation pipeline.  

\noindent \textbf{Fact-based Visual Question Answering (FVQA)} \cite{fvqa}: Each data point in this dataset consists of an image, a question about the image, the answer, and a fact supporting the answer. This dataset has also provided an unstructured knowledge source containing all the facts (i.e., sentences) we need to answer the question about the image. We use 70\% of samples (4077) in this dataset for train set and augment them similar to \citet{okvqa-passage-retrieval} with five hard negatives retrieved by BM25, 15\% for validation (874), and 15\% for test set (874). We use the original FVQA dataset to evaluate the performance of our end-to-end retrieval and answer generation pipeline.


\subsection{Experimental Setup}

\subsubsection*{\textbf{Retriever Training Setup}}
In our experiments, we use the Adam optimizer \cite{adam} with a batch size of 16 and a learning rate of $10^{-5}$. We use a linear learning rate scheduler with 10\% of total steps as warm-up steps. We also use gradient clipping with the value of 1.0. The maximum input length of each encoder is set to 400 tokens. We train each model for 2 epochs on OK-VQA and 4 epochs on FVQA. All the experiments are conducted on a machine with a Nvidia RTX8000 GPU with 49GB of memory and 256GB of RAM. We use Faiss \cite{faiss} to index the learned embeddings with a flat index for efficient dense retrieval. For BM25, Pyserini is used.




\subsubsection*{\textbf{Multi-Modal Fusion-in-Decoder Training Setup}}
We use the AdamW optimizer with a batch size of 1 with 32 gradient accumulation steps, which results in an effective batch size of 32. We utilize a learning rate of $5\times10^{-5}$ and weight decay of $0.1$ for training the MM-FiD model. Given the training dataset sizes, we use a linear learning rate scheduler with 800 and 200 warm-up steps for the OK-VQA and FVQA datasets, respectively. We train the model for 5000 (OK-VQA) and 2000 (FVQA) gradient update steps. We create a checkpoint of the model every 500 training steps and select the best checkpoint based on its performance on the validation set. We also use gradient clipping at 1.0 for training. The maximum encoder's input length of each question and passage pair is set to 420 (OK-VQA) and 64 (FVQA) tokens. 
Since the answers in both datasets are short, we set the MM-FiD's output length to $16$. MM-FiD's decoder uses beam search \cite{beam-search} with a beam size of 2 for answer generation. We train MM-FiD using $32$ (OK-VQA) and $5$ (FVQA) supporting passages for each question and image pair, where the supporting passages are retrieved using the proposed \drframework model. The MM-FiD experiments are conducted on a machine with a single Nvidia RTX8000 GPU with 49GB of memory and 128GB of RAM. Following \cite{fvqa}, we use five-fold cross-validation for the FVQA dataset. Note that we train an individual \drframework for each fold to avoid data leaks.

\begin{table*}
    \centering
    \caption{Passage retrieval performance for KI-VQA tasks on OK-VQA and FVQA datasets. The superscript $^*$ denotes statistically significant improvement compared to all the baselines based on two-tailed paired t-test with Bonferroni correction ($p < 0.05$).}
    \vspace{-0.2cm}
    \resizebox{\textwidth}{!}{
    \begin{tabular}{l|cc|cc|cc|cc}
     \textbf{Dataset} & \multicolumn{4}{c}{\textbf{OK-VQA}} & \multicolumn{4}{c}{\textbf{FVQA}}\\
    \hline
    \multirow{2}{*}{\textbf{Model}} & \multicolumn{2}{c}{\textbf{Validation}} & \multicolumn{2}{c}{\textbf{Test}} & \multicolumn{2}{c}{\textbf{Validation}} & \multicolumn{2}{c}{\textbf{Test}} \\
    & \textbf{MRR@5} & \textbf{P@5} & \textbf{MRR@5} & \textbf{P@5} & \textbf{MRR@5} & \textbf{P@1} & \textbf{MRR@5} & \textbf{P@1}\\
    \hline
    \multicolumn{9}{c}{Sparse Retrievers} \\
    \hline
    BM25  & 0.2565 & 0.1772 & 0.2637 & 0.1755 & 0.3368 & 0.2700 & 0.3509 & 0.2848 \\
    BM25-Obj (CombMax)  & 0.3772 & 0.2667 & 0.3686& 0.2541 & 0.3903 & 0.3272 & 0.4057 & 0.3421 \\
    \hline
    \multicolumn{9}{c}{Symmetric Single-Encoding Dense Retrievers} \\
    \hline
    Dense-BERT (question, passage)  & 0.4555& 0.3155 & 0.4325 & 0.3058 & 0.3860 & 0.3089 & 0.3836& 0.3020 \\ 
    Dense-BERT (question + caption, passage) $\rightarrow$ Our $E_T$  & 0.5843 & 0.4445 & 0.5797& 0.4420 & 0.4409& 0.3558 & 0.4292 & 0.3409 \\ 
    Dense-LXMERT $\rightarrow$ Our $E_{MM}$ & 0.5722 & 0.4276 & 0.5465 & 0.4066 & 0.4293 & 0.3478 & 0.4269 & 0.3409 \\ \hline
    \multicolumn{9}{c}{Asymmetric Dual-Encoding Dense Retrievers} \\\hline
    BERT-LXMERT & 0.4704& 0.3364 & 0.4526& 0.3329 & 0.1455 & 0.1006 & 0.1477 & 0.1029 \\
    \hline
    \multicolumn{9}{c}{Symmetric Dual-Encoding Dense Retrievers} \\
    \hline
    \drframework & \textbf{0.6260$^*$} & \textbf{0.4890$^*$} & \textbf{0.6469$^*$} & \textbf{0.5059$^*$} & \textbf{0.5833$^*$} & \textbf{0.4931$^*$} & \textbf{0.5618$^*$} & \textbf{0.4713$^*$} \\
    \quad \% relative improvement w.r.t. the best baseline & \textcolor{teal}{\textbf{7.1\% $\uparrow$}} & \textcolor{teal}{\textbf{10.0\% $\uparrow$}} & \textcolor{teal}{\textbf{11.6\% $\uparrow$}} & \textcolor{teal}{\textbf{14.5\% $\uparrow$}} & \textcolor{teal}{\textbf{32.3\% $\uparrow$}} & \textcolor{teal}{\textbf{38.6\% $\uparrow$}} & \textcolor{teal}{\textbf{30.9\% $\uparrow$}} & \textcolor{teal}{\textbf{37.8\% $\uparrow$}} \\
    \end{tabular}
    }
    \label{tab:retrieval_results}
    \vspace{-0.3cm}
\end{table*}

\subsubsection*{\textbf{Evaluation Metrics}}
To be consistent with the literature, we use the common metrics suggested for each dataset. Following \citet{okvqa-passage-retrieval}, we use mean reciprocal rank (MRR) and precision of the top five retrieved documents (MRR@5 and P@5) for evaluating the retrieval models on the OK-VQA dataset. We use MRR@5 and P@1 for retrieval evaluation on the FVQA dataset. Since the FVQA dataset provides only one ground truth passage per question, we use Precision@1 as the evaluation metric. We use a two-tailed paired t-test with Bonferroni correction to identify statistically significant improvements (p-value $< 0.05$).

For the evaluation of the answer generation model for the OK-VQA dataset, we follow the official evaluation script provided by \citet{okvqa}. It uses the evaluation metric for VQA \cite{VQA} task, which relies on human annotations. For evaluation on the FVQA dataset, we follow \citet{fvqa} and use Top-1 Accuracy or Exact Match (EM) as the evaluation metric, in which we lowercase answers and remove articles (e.g. a, an, the) and punctuation.


\subsection{Passage Retrieval Results for KI-VQA Tasks}
\label{sec:passage-ret-results}

\subsubsection*{\textbf{Baselines.}} We compare the proposed dense retrieval framework with the following baselines: 
\begin{itemize}[leftmargin=*]
    \item \textbf{Sparse (Term Matching) Retrieval Models:} We use two sparse retrieval baselines: (1) \textbf{BM25:} this baseline uses the BM25 formulation \cite{Robertson1995OkapiBM25} with questions as queries, ignoring the images, and passages as documents. (2) \textbf{BM25-Obj (CombMax):} this approach extracts 36 objects from the image (objects are generated by a Faster R-CNN \cite{faster-r-cnn} model pre-trained on Visual Genome \cite{Anderson2018BottomUpAT,Krishna2016VisualGC}) and concatenates each object's name to the question as the query and uses the BM25 formulation to retrieve passages. Then it uses CombMax \cite{combmax1, combmax2} to aggregate these 36 ranked lists. \citet{okvqa-passage-retrieval} explored other rank aggregation approaches as well and CombMax was found to be the best solution for this task. 

    \item \textbf{Symmetric Single-Encoding Dense Retrieval Models:} We use three baselines in this category: (3, 4) \textbf{Dense-BERT:} a BERT-based dense retrieval model (similar to DPR \cite{dpr}) with the same training objective as ours. We provide the results for two variations of this model, an image-independent approach whose query encoder only encodes the question, and an image-dependent approach whose query encoder takes the concatenation of the question and the image caption (captions are generated by ExpansionNet~v2 \cite{expansionv2}). The latter is the same as our $E_T$ encoder. (5) Dense-LXMERT is a model that uses a multi-modal encoder, i.e., LXMERT, to encode queries and passages. It uses masked image tokens on the passage side. This approach is our $E_{MM}$ encoder. 

    \item \textbf{Asymmetric Dual-Encoding Dense Retrieval Models:} In this category, we use BERT-LXMERT, proposed in \cite{okvqa-passage-retrieval}, that uses BERT for passage encoding and LXMERT for query encoding.\footnote{The original paper \cite{okvqa-passage-retrieval} refers to this baseline as Dense-LXMERT. We rename it to BERT-LXMERT to avoid confusion.} 
\end{itemize}

For fair comparison, we use the same training and evaluation process for all (our and baseline) models. To the best of our knowledge, our baseline results are the highest reported in the literature. 



\subsubsection*{\textbf{Comparison Against Retrieval Baselines.}}
\begin{figure}[t]
    \centering
    \includegraphics[width=\linewidth]{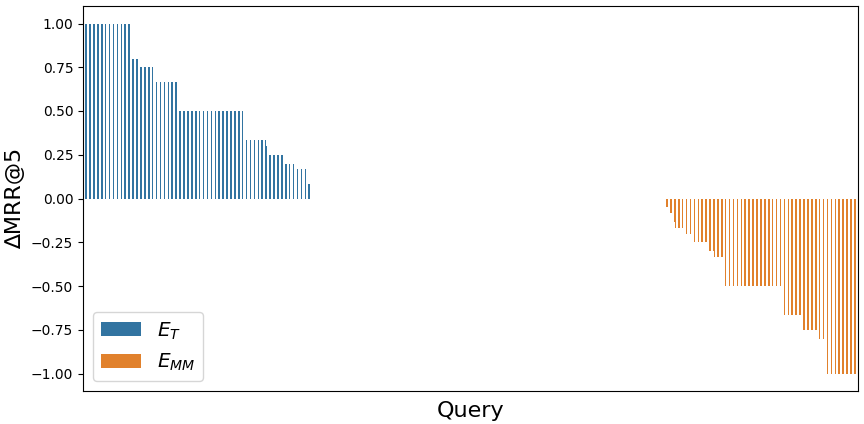}
    \vspace{-0.65cm}
    \caption{Difference between reciprocal rank (RR) obtained by $E_T$ and $E_{MM}$ for each query on the OK-VQA test set. The blue / orange color denotes the queries where $E_T$ / $E_{MM}$ wins.}
    \label{fig:mrr-diff-lxmert-captbert}
    \vspace{-0.3cm}
\end{figure}

The passage retrieval results are reported in Table \ref{tab:retrieval_results}. We observe that dense retrieval models generally outperform sparse retrieval baselines, confirming our design choice to focus on dense retrieval for KI-VQA tasks. 
Interestingly, sparse retrieval models achieve higher MRR on FVQA than on OK-VQA, while dense retrieval models perform significantly better on the OK-VQA dataset. This observation suggests that relevant documents in FVQA perhaps have higher term overlap with the questions and image objects. The results show that image-independent models (i.e., BM25 and Dense-BERT that only encodes the question) underperform their own variant with image information (i.e., generated objects or captions). This highlights the importance of representing images in KI-VQA tasks. Furthermore, Table \ref{tab:retrieval_results} shows that the asymmetric dense retrieval baseline (BERT-LXMERT) does not perform as well as symmetric dense retrieval baselines. In particular, BERT-LXMERT shows a poor performance on the FVQA dataset. That can be due to the smaller training set in FVQA,\footnote{FVQA not only has fewer training questions, but only has a single relevant passage per question.} as asymmetric models often require more training data.  Another observation from these results is that performance on validation and test sets are relatively close, therefore it is safe to argue that performances on the validation sets are generalizable to the test sets and no aggressive overfitting is observed. The results show that our own encoders $E_T$ and $E_{MM}$ are the best performing baselines. We conducted an experiment to see if these two encoding approaches provide complementary information. To this aim, we compute the reciprocal rank (RR) obtained by $E_T$ and $E_{MM}$ for each query in OK-VQA and plot their differences in Figure~\ref{fig:mrr-diff-lxmert-captbert}. For better visualization, this figure sorts the queries with respect to their $\Delta MRR$ in descending order. Figure~\ref{fig:mrr-diff-lxmert-captbert} shows that $E_T$ and $E_{MM}$ perform similarly for about half of the queries, while $E_T$ performs better for about $25\%$ of queries and $E_{MM}$ performs better for the other $\sim25\%$. \textbf{This shows that these two encoders contain complementary information and their aggregation can improve the results -- confirming the motivation of designing \drframework.}

\begin{table*}
    \centering
    \caption{Ablation study for iterative knowledge distillation in \drframework. The superscript $^*$ denotes statistically significant improvement compared to all the ablation cases based on two-tailed paired t-test with Bonferroni correction ($p < 0.05$).}
    \vspace{-0.3cm}
    \begin{tabular}{l|cc|cc|cc|cc}
     \textbf{Dataset} & \multicolumn{4}{c}{\textbf{OK-VQA}} & \multicolumn{4}{c}{\textbf{FVQA}}\\
    \hline
    \multirow{2}{*}{\textbf{Model}} & \multicolumn{2}{c}{\textbf{Validation}} & \multicolumn{2}{c}{\textbf{Test}} & \multicolumn{2}{c}{\textbf{Validation}} & \multicolumn{2}{c}{\textbf{Test}} \\
    & \textbf{MRR@5} & \textbf{P@5} & \textbf{MRR@5} & \textbf{P@5} & \textbf{MRR@5} & \textbf{P@1} & \textbf{MRR@5} & \textbf{P@1}\\
    \hline
   
    \drframework (joint encoders, no KD) & 0.5930 & 0.4507 & 0.5405 & 0.4263 & 0.4669 & 0.3764 & 0.4498 & 0.3684 \\
    \drframework (isolated encoders, no KD) & 0.6236 & 0.4771 & 0.6330 & 0.4972 & 0.5551 & 0.4668 & 0.5313 & 0.4439 \\
    \drframework & \textbf{0.6260} & \textbf{0.4890$^*$} & \textbf{0.6469$^*$} & \textbf{0.5059$^*$} & \textbf{0.5833$^*$} & \textbf{0.4931$^*$} & \textbf{0.5618$^*$} & \textbf{0.4713$^*$} \\
    \end{tabular}
    \label{tab:retrieval_ablation}
    \vspace{-0.3cm}
\end{table*}

Results obtained by \drframework suggest statistically significant improvements compared to all the baselines. We achieve 11.6\% higher MRR and $14.5\%$ higher P@5 than the best baseline (including our own encoders $E_T$ and $E_{MM}$) on the OK-VQA test set and  30.9\% higher MRR and $37.8\%$ higher P@1 on the FVQA test set. We believe that our symmetric dual encoding approach works well without requiring a large scale training set, which justifies substantially larger gain on the FVQA dataset. Note that \drframework also uses BERT and LXMERT and the obtained improvements are not due to larger model parameters or different pretraining. However, compared to Dense-BERT and Dense-LXMERT, \drframework has the ability to take advantage of knowledge from both uni- and multi-modal language models. BERT-LXMERT, however, could not take advantage of such extra implicit ``knowledge'' effectively due to its asymmetric design.

\begin{figure}
    \centering
    \includegraphics[width=\linewidth]{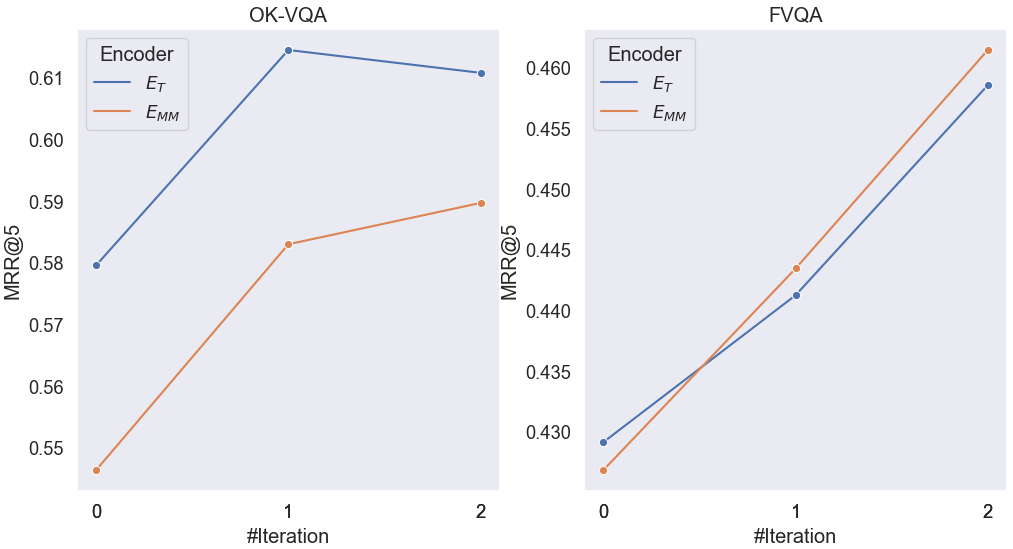}
    \vspace{-0.7cm}
    \caption{\drframework performance at different iterations of the proposed iterative knowledge distillation approach on the test set of both OK-VQA and FVQA datasets.}
    \label{fig:iterative-kd}
    \vspace{-0.5cm}
\end{figure}

\subsubsection*{\textbf{\drframework Ablations.}} To empirically study the impact of each decision we made in designing \drframework, we report ablation results in Table~\ref{tab:retrieval_ablation}. The first row of the table is associated with a model that jointly used both encoders ($E_T$ and $E_{MM}$) at both training and evaluation. The second row, on the other hand, train each encoder separately until convergence and only concatenates them at inference. There is no knowledge distillation in either of these two models. The results show that \drframework outperforms both of these models, demonstrating the effectiveness of the proposed iterative knowledge distillation approach for dual encoding. The improvements are statistically significant in nearly all cases, except for MRR@5 on the OK-VQA validation set. We further plot the retrieval performance at each knowledge distillation training step in Figure~\ref{fig:iterative-kd}. We observe that in the first three iterations the performance of both encoders generally increases on both datasets. Models show different behavior on the different datasets in Figure~\ref{fig:iterative-kd}. This shows that the number of iterations in the knowledge distillation approach is dataset-dependent and should be tuned for best results.

\begin{table*}
    \centering
    \caption{Question answering performance for models with explicit knowledge source on both OK-VQA and FVQA datasets. All these models are relatively comparable in terms of model size and contain less than 1 billion parameters.}
    \vspace{-0.3cm}
    \resizebox{\textwidth}{!}{
    \begin{tabular}{l|ll||l|ll}
         \multicolumn{3}{c||}{\textbf{OK-VQA}} & \multicolumn{3}{c}{\textbf{FVQA}}\\\hline
        \textbf{Model} & \textbf{Knowledge Source} & \textbf{Acc.} & \textbf{Model} & \textbf{Knowledge Source} & \textbf{Top-1 Acc.}  \\
        \hline
        KAT-base \cite{kat} & WikiData & 40.93 & BAN \cite{ban} & None &  35.69 \\
        RVL \cite{rvl} & ConceptNet & 39.04 & RVL \cite{rvl} & ConceptNet  & 54.27 \\
        MAVEx \cite{mavex} & Wikipedia, ConceptNet, Google Images & 40.28 & BAN \cite{ban} + KG-AUG \cite{Li2020BoostingVQ} & None &  38.58 \\
        UnifER+ViLT \cite{unifer} & ConceptNet & 42.13 & UnifER + ViLT \cite{unifer} & FVQA Corpus & 55.04 \\
        VRR \cite{vrr} & Google Search & 39.20  & Top1-QQmaping \cite{fvqa} & FVQA Corpus & 52.56 \\
        LaKo-base \cite{lako} & ConceptNet, DBPedia, WebChild & 42.21 & Top3-QQmaping \cite{fvqa} & FVQA Corpus & 56.91 \\
        \hline
        \mmfid & None & 42.82 & \mmfid  & None & 52.78 \\
        \drframework+\mmfid & Wikipedia & \textbf{44.57} &\drframework+\mmfid & FVQA Corpus & \textbf{61.80} \\
        \multicolumn{2}{l}{\quad \% relative improvement w.r.t. the best baseline} & \textcolor{teal}{\textbf{5.5\% $\uparrow$}} & \multicolumn{2}{l}{\quad \% relative improvement w.r.t. the best baseline} & \textcolor{teal}{\textbf{8.5\% $\uparrow$}} \\
    \end{tabular}
    }
    \label{tab:mmfid_results}
    \vspace{-0.3cm}
\end{table*}

\subsection{Question Answering Results for KI-VQA}
In this section, we report and discuss the end-to-end retrieval and answer generation results. A wide range of question answering methods has been applied to KI-VQA tasks. Not all of these methods are publicly available and not all of them use the same knowledge source. We compare our methods to the best performing models in the literature with relatively similar model size. Note that \mmfid contains 220 million parameters. In our first set of experiments, we also exclude the models that use GPT-3 (175 billion parameters) as an implicit ``knowledge'' source. The results are reported in Table~\ref{tab:mmfid_results}. As mentioned in the table, different approaches on OK-VQA use different knowledge sources, such as ConceptNet, Wikidata, Wikipedia, Google Search, and Google Images. The FVQA dataset released a fact corpus which is used by several models. We observe that \mmfid that uses passages retrieved by \drframework for answer generation outperforms all the baselines listed in Table~\ref{tab:mmfid_results}. We observe 8.5\% improvements on FVQA compared to the best performing baseline. This table also includes the results for \mmfid without any supporting document (i.e., without retrieval). We observe that the \mmfid is able to produce competitive performance even without utilizing retrieval results. The reason is that these large language models contain a lot of information in their parameters from pre-training phase and they can answer many questions based on their internal implicit ``knowledge''. \mmfid without retrieval even outperforms all the baselines on OK-VQA. Note that the number of parameters in \mmfid (220M) is comparable to the baselines. The results suggest that employing retrieval results leads to larger gain on FVQA than on OK-VQA, possibly due to the nature of the questions in fact-based visual question answering.

\begin{table}
    \centering
    \caption{QA performance of SOTA models that rely on the output of GPT-3 as a ``knowledge source'' on OK-VQA.}
    \vspace{-0.3cm}
    \begin{tabular}{p{2.5cm}|lc}
        \textbf{Model}  & \textbf{Knowledge source} & \textbf{Accuracy} \\
        \hline
        PICa-full \cite{pica} & Frozen GPT-3  & 48.00 \\
        KAT-base \cite{kat} & Frozen GPT-3, Wikidata & 50.58 \\
        \hline
        \drframework + \mmfid & Frozen GPT-3, Wikipedia & \textbf{51.02} \\
    \end{tabular}
    \label{tab:gpt3_results}
\end{table}

Note that some models use the outputs produced by GPT-3 as a knowledge source and often outperform those models above that use explicit knowledge. However, it is difficult to draw conclusions, as the GPT-3 training set is unknown and it has 175B parameters, making it extremely expensive to run and not truly comparable to any other model mentioned above in terms of capacity. That being said, we do compare our model against state-of-the-art baselines with comparable model size that use GPT-3's output as supporting evidence. The results on the OK-VQA dataset are reported in Table~\ref{tab:gpt3_results}.\footnote{We do not have access to the GPT-3's output for FVQA questions.} When our model uses GPT-3's output in addition to passages, it still outperforms its alternatives, but with a smaller margin, highlighting the impact of document quality in KI-VQA tasks.

\begin{table}
    \centering
    \caption{Ablation study of the the proposed KI-VQA pipeline. The superscript $^*$ denotes statistically significant improvement compared to all the ablation cases based on two-tailed paired t-test with Bonferroni correction ($p < 0.05$).}
    \vspace{-0.3cm}
    \resizebox{\linewidth}{!}{
    \begin{tabular}{ll|cc}
        \multirow{2}{*}{\textbf{Retriever}} & \textbf{Answer} & \textbf{OK-VQA} & \textbf{FVQA} \\
         & \textbf{Generator} & \textbf{Acc} & \textbf{Top-1 Acc} \\
        \hline
        \drframework & FiD & 39.48 & 60.85 \\\hline
        BM25-Obj (CombMax) \cite{okvqa-passage-retrieval} & MM-FiD & 41.97 & 54.65 \\ 
        Best retrieval baseline from Table~\ref{tab:retrieval_results} & MM-FiD & 41.82 & 52.78 \\
        \drframework & MM-FiD & \textbf{44.57}$^*$ & \textbf{61.80}$^*$ \\ 
    \end{tabular}
    }
    \label{tab:mmfid-retrievers}
    \vspace{-0.5cm}
\end{table}

\subsubsection*{\textbf{Question Answering Ablations and Analysis.}} For a deeper understanding of the proposed answer generation solution, we conduct careful ablation studies whose results are reported in Table~\ref{tab:mmfid-retrievers}. The results on both datasets suggest that when using the same retriever (i.e., \drframework), \mmfid outperforms its uni-modal variation, FiD \cite{fid}, that has been also used for KI-VQA tasks, for example in KAT \cite{kat}. 
Moreover, Table~\ref{tab:mmfid-retrievers} demonstrates the impact of different retrieval models on the final answer generation. For example, using \drframework for retrieval instead of the best performing retrieval baseline from Table~\ref{tab:retrieval_results} would lead to 13\% higher accuracy in FVQA, highlighting the importance of retrieval in the KI-VQA pipeline.

\begin{figure}
    \centering
    \includegraphics[scale=0.26]{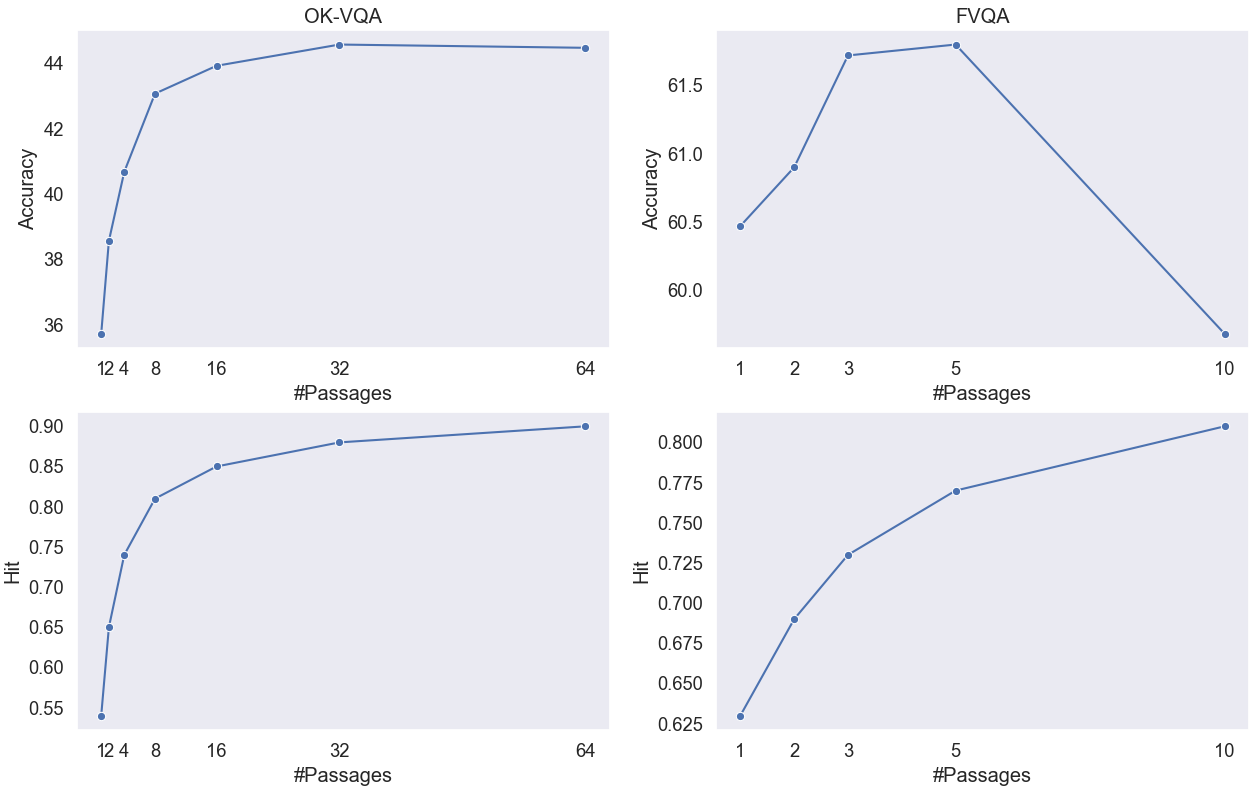}
    \vspace{-0.3cm}
    \caption{\mmfid accuracy and the hit ratio in the supporting passages at different ranking cut-off levels.}
    \label{fig:mmfid-acc-ctx-size}
    \vspace{-0.4cm}
\end{figure}

Figure~\ref{fig:mmfid-acc-ctx-size} plots the sensitivity of \mmfid performance to the number of passages retrieved by \drframework. It also plots the hit ratio, i.e., the ratio of success retrieving at least one relevant passage to be presented to \mmfid. Generally speaking, the more documents we feed to \mmfid on OK-VQA, the higher the question answering accuracy. Its accuracy becomes relatively stable after retrieving 16 documents. The accuracy curve follows the same behavior as the hit curve on OK-VQA. However, FVQA demonstrates a substantially different behavior. The highest accuracy is achieved when only five supporting passages are retrieved. That is due to the nature of the dataset, where there is only one relevant fact for each question and retrieving more (potentially inaccurate) facts may confuse the answer generation model. Note that \drframework reaches a hit ratio of 70\% by only retrieving two passages on FVQA, while the same model needs to retrieve 4 passages to reach the same level of hit ratio on OK-VQA. This finding suggest that it is worth studying automatic prediction of ranking cut-off for KI-VQA tasks in the future.

\section{Conclusions and Future Work}
This paper presented \drframework, a novel symmetric dense retrieval framework based on dual uni-modal and multi-modal encoding. We propose an iterative knowledge distillation approach for updating these two encoding representation spaces and aggregating them at inference. It also proposed \mmfid, an extension to the fusion-in-decoder architecture \cite{fid} for multi-modal data. Extensive experiments on two well-established datasets, OK-VQA and FVQA, suggested that retrieving passages using \drframework and using them to generate answers via \mmfid substantially outperforms state-of-the-art baselines with comparable capacity. For instance, this approach led to 37.8\% retrieval improvement in terms of P@1 and 8.5\% exact match accuracy improvement on FVQA test set compared to the best performing baselines. We demonstrated the impact of every design decision we made in both \drframework and \mmfid through extensive ablation studies and highlight open areas for future explorations. For example, our results suggest accurate prediction of `when to retrieve' is an impactful area for KI-VQA tasks. Hence, exploring retrieval performance prediction and ranking cut-off truncation in KI-VQA tasks can potentially be a fruitful future direction. We also intend to explore universal knowledge-intensive models for both textual and multi-modal inputs. We further plan to expand the applications of knowledge-intensive multi-modal tasks beyond question answering.

\section*{Acknowledgment}

This work was supported in part by the Center for Intelligent Information Retrieval, in part by NSF grant \#2106282, and in part by Lowe's. Any opinions, findings and conclusions or recommendations expressed in this material are those of the authors and do not necessarily reflect those of the sponsor.

\bibliographystyle{ACM-Reference-Format}
\balance
\bibliography{00-main}

\end{document}